# RePAD: Real-time Proactive Anomaly Detection for Time Series

Ming-Chang Lee[1], Jia-Chun Lin[2], and Ernst Gunnar Gran[3]

*[1,2,3]Department of Information Security and Communication Technology, Norwegian University of Science and Technology,*
*Ametyst-bygget, 2815 Gjøvik, Norway*
*[3]Simula Research Laboratory, 1364 Fornebu, Norway*

*[1] ming-chang.lee@ntnu.no*
*[2]jia-chun.lin@ntnu.no*
*[3]ernst.g.gran@ntnu.no*

# RePAD: Real-time Proactive Anomaly Detection for Time Series

Ming-Chang Lee[1], Jia-Chun Lin[2], and Ernst Gunnar Gran[3]
[1,2,3] Department of Information Security and Communication Technology, Norwegian University of Science and Technology, 2815 Gjøvik, Norway
[3]Simula Research Laboratory, 1364 Fornebu, Norway
[1] ming-chang.lee@ntnu.no
[2] jia-chun.lin@ntnu.no
[3] ernst.g.gran@ntnu.no

**Abstract.** During the past decade, many anomaly detection approaches have been introduced in different fields such as network monitoring, fraud detection, and intrusion detection. However, they require understanding of data pattern and often need a long off-line period to build a model or network for the target data. Providing real-time and proactive anomaly detection for streaming time series without human intervention and domain knowledge is highly valuable since it greatly reduces human effort and enables appropriate countermeasures to be undertaken before a disastrous damage, failure, or other harmful event occurs. However, this issue has not been well studied yet. To address it, this paper proposes RePAD, which is a Real-time Proactive Anomaly Detection algorithm for streaming time series based on Long Short-Term Memory (LSTM). RePAD utilizes short-term historical data points to predict and determine whether or not the upcoming data point is a sign that an anomaly is likely to happen in the near future. By dynamically adjusting the detection threshold over time, RePAD is able to tolerate minor pattern change in time series and detect anomalies either proactively or on time. Experiments based on two time series datasets collected from the Numenta Anomaly Benchmark demonstrate that RePAD is able to proactively detect anomalies and provide early warnings in real time without human intervention and domain knowledge.

## 1 Introduction

Streaming time series is generated everywhere from stock exchange, smart meters, smart homes, and system monitoring tools to medical devices implanted within the human body. Regardless of their sources, a time series is a series of data points that are generated continuously and evenly indexed in time order. This kind of data already constitutes a large portion of the world's data and we expect that the amount of time series data will keep increasing due to the explosive growth of diverse Internet of Things devices.

Anomaly detection refers to any approach or technique designed to find anomalies in data. During the past decade, many approaches and methods have been introduced for time series anomaly detections in different domains such as intrusion detection, fraud detection, system failure detection, CPU utilization monitoring, smart homes, healthcare, etc. Examples include [2-8]. However, most approaches require either domain knowledge (such as a set of pre-labeled training data or a statistical model) or human intervention (such as collecting sufficient and representative training data, conducting an off-line learning/training process, tuning appropriate parameters, etc.), consequently limiting their applicability. To provide real-time and proactive anomaly detection with neither human intervention nor domain knowledge could be highly valuable in many different domains and applications; such an approach can be easily and immediately applied to time series data and it enables appropriate actions and countermeasures to be taken before the occurrence of a catastrophic failure/event.

To facilitate the above-mentioned approach, this paper proposes a Real-time and Proactive Anomaly Detection algorithm (RePAD for short) for streaming time series based on Long Short-Term Memory (LSTM)[23]. LSTM is a special type of recurrent neural networks, superior in capturing short-term dependencies when making prediction over sequence data [9]. Previous work [10][11][12][22] show that LSTM provides better prediction accuracy than many other approaches and neural networks. It is well known that more training time will be required if an LSTM model have more training data, a more complex structure, etc. In order to take advantage of the superior prediction performance of LSTM and meanwhile achieve real-time and proactive anomaly detection, RePAD utilizes short-term historical data points to predict and determine whether the upcoming data point is a sign that an anomaly is likely to happen in the near future or not. Due to the small size of training data, a simple LSTM model is sufficient and can be trained in a short period of time. In addition, in order to be able to tolerate minor pattern changes in time series and detect anomalies proactively, RePAD dynamically adjusts its threshold to detect anomalies, and decides if a new LSTM model needs to be retrained for accommodating changing patterns or not. In other words, RePAD performs LSTM retraining only when it is necessary. This feature makes RePAD adaptive and able to offer real-time anomaly detection.

To demonstrate the proactive anomaly detection performance of RePAD, we chose two real-world time series datasets from the Numenta Anomaly Benchmark (NAB) [1] and compare RePAD with two anomaly detection approaches provided by Twitter [2]. The experiment results show that RePAD is able to detect anomalies either proactively or on time without any human intervention or domain knowledge.

The rest of the paper is organized as follows: Sections 2 and 3 describe LSTM configuration and related work, respectively. In Section 4, we introduce the details of RePAD. Section 5 presents and discusses the experiments and the corresponding results. In Section 6, we conclude this paper and outline future work.

## 2 LSTM Configuration

LSTM was proposed and designed to learn long short-term dependencies and model temporal sequences [23]. The network structure of LSTM is similar to that of a recurrent neural network, except that the nonlinear units in each hidden layer of LSTM are memory blocks. Each memory block has its own memory cells and three gates: input, output, and forget. The input gate decides to store data in the memory cell. The output gate determines if current memory content should be output. The forget gate decides if the current memory content should be erased or not. These features enable LSTM to address the vanishing gradient problem [13].

It is well-known that if an LSTM model has a complicated structure or the training data is large, the required training time will significantly increase. To achieve real-time anomaly detection, the LSTM models utilized and (re)trained by RePAD always have a simple structure and a fast learning speed: only one hidden layer, 10 hidden units in the hidden layer, and a learning rate of 0.15. With respect to epoch (which is defined as one forward pass and one backward pass of all the training data), it is clear that too many epochs might overfit the training data, whereas too few epochs may underfit the training data. To address this issue, whenever RePAD needs to train/retrain LSTM, Early Stopping [14] is employed, which is an approach to automatically determine the number of epochs for preventing LSTM from overfitting the training data. In this paper, Early Stopping always chooses a number between 1 and 50.

## 3 Related Work

Anomaly detection approaches developed during the past decades can be classified into two categories: Statistical-learning approaches and machine-learning approaches. The former category of approaches fit a statistical model to a given normal dataset and then uses the model to determine whether a new data point fits this model or not. If the data point has a low probability to be generated from the model, the data point is considered to represent an anomaly. For example, Twitter introduced two anomaly detection algorithms based on statistical learning and Seasonal Hybrid ESD (S-H-ESD) [2]: AnomalyDetectionTs (ADT for short) and AnomalyDetectionVec (ADV for short). Both of them have been implemented and included in an open-source R package [15]. ADT is designed to detect one or more statistically significant anomalies in a given time series, while ADV is designed to detect one or more statistically significant anomalies in a given vector of observations without timestamp information. Note that in ADV, the period parameter must be pre-configured such that ADV can carry out its anomaly detection on a periodic basis.

Previous machine-learning approaches used in anomaly detection can be further divided into supervised learning and unsupervised learning. The former type of approaches require a set of labeled training data in which each data point/instance must be labeled as normal or anomalous. Examples of using this approach include [4] and [5]. However, it is expensive and difficult to obtain sufficient and accurately labeled training data since labeling is often done by human experts manually.

On the contrary, unsupervised learning approaches do not require pre-labeled training data, but they require a long off-line period to train their neural networks. Wu et al. [6] proposed Hierarchical Temporal Memory (HTM) to capture changing patterns in time series. However, HTM requires 15% of the whole dataset to be non-anomalous so that it can use this data to train its neural network. Lee et al. [9] proposed a zero-positive anomaly detection algorithm called Greenhouse for time series anomaly detection based on LSTM. Greenhouse requires three sets of non-anomalous training data to train and test its LSTM model. During the training phase, Greenhouse adopts a Look-Back and Predict-Forward strategy to detect anomalies. For a given time point $t$, a window of length $B$ of most recently observed values is used as "Look-Back" to predict a subsequent window of values of length *F* as “Predict-Forward”. This feature enables Greenhouse to adapt to pattern changes in the training data. However, if the training data is not representative, Greenhouse might not be able to capture pattern changes in real-world time series. Other examples based on semi-supervised learning include [7][8].

Similar to Greenhouse, our proposed RePAD also follows the Look-Back and Predict-Forward strategy. However, different from Greenhouse, RePAD employs this strategy to do on-line LSTM training and real-time anomaly detection. In other words, RePAD does not need to pre-train its LSTM model, meaning that RePAD is a completely unsupervised learning approach. The details will be introduced in the next section.

## 4 The details of RePAD

As mentioned earlier, RePAD aims to proactively detect anomalies in streaming time series. To achieve this goal, RePAD employs a lightweight LSTM model to predict the values of the data points arriving at the next $f$ time points, based on the data values observed at the past $b$ continuous time points. In this paper, $b$ is called the Look-Back parameter, whereas $f$ is called the Predict-Forward parameter. Note that both of them are small integers, and that the $b$ time points and the $f$ time points do not need to be adjacent.

In order to explain how RePAD works and make the explanation easy to follow, let us assume that $b = 3$ and $f = 1$. Fig. 1 illustrates the algorithm of RePAD. Let $t$ be the current time point, and $t$ starts from 0. When RePAD is initialized and launched, it requires to firstly collect data points $v_0$ and $v_1$ at time points 0 and 1, respectively. When the current time point advances to 2 (i.e., $t = 2$), RePAD is able to train an LSTM model by taking the three observed data points $[v_0, v_1, v_2]$ as the training data. The trained LSTM model, denoted by $M$, is then immediately used to predict the value of the next data point, denoted by $\widehat{v_3}$. When $t = 3$, RePAD retrains a new LSTM model with the most recent $b$ data points, i.e., $[v_1, v_2, v_3]$. This new LSTM model replaces $M$ to predict the value of the next data point, denoted by $\widehat{v_4}$. Similarly, when $t = 4$, RePAD retrains another new LSTM model with data points $[v_2, v_3, v_4]$ to replace $M$, and then uses this new model to predict $\widehat{v_5}$. When $t = 5$ (i.e., $t = 2b - 1$), RePAD is able to calculate $AARE_5$ based on Equation 1.

$$AARE_t = \frac{1}{b} \cdot \sum_{y=t-b+1}^{t} \frac{\left|v_y - \widehat{v_y}\right|}{v_y} \, , t \geq 2b - 1 \tag{1}$$

where $v_y$ is the observed data value at time point $y$, and $\widehat{v_y}$ is the forecast data value at $y$. Since we assume that $b = 3$, $AARE_5 = \frac{1}{3} \cdot \sum_{y=3}^{5} \frac{\left|v_y - \widehat{v_y}\right|}{v_y} = \frac{1}{3} \cdot \left( \frac{|v_3 - \widehat{v_3}|}{v_3} + \frac{|v_4 - \widehat{v_4}|}{v_4} + \frac{|v_5 - \widehat{v_5}|}{v_5} \right)$. Note that AARE stands for Average Absolute Relative Error, which is a well-known measure for determining the prediction accuracy of a forecast approach [16]. A low AARE value indicates that the predicted values are close to the observed values.

When $t = 6$ (i.e., $t = 2b$), RePAD is able to calculate $AARE_6$ based on Equation 1, i.e., $AARE_6 = \frac{1}{3} \cdot \sum_{y=4}^{6} \frac{|v_y - \widehat{v_y}|}{v_y} = \frac{1}{3} \cdot \left( \frac{|v_4 - \widehat{v_4}|}{v_4} + \frac{|v_5 - \widehat{v_5}|}{v_5} + \frac{|v_6 - \widehat{v_6}|}{v_6} \right)$.

**RePAD algorithm**
**Input**: Data points in a time series
**Output**: Anomaly notifications
**Procedure:**

1: Let $t$ be the current time point and $t$ starts from 0; Let *flag* be True;
2: **While** time has advanced {
3: Collect data point $v_t$;
4: **if** $t \geq b - 1$ and $t < 2b - 1$ { //i.e., $2 \leq t < 5$, if $b = 3$
5: Train an LSTM model by taking $[v_{t-b+1}, v_{t-b+2} \dots, v_t]$ as the training data;
6: Let $M$ be the resulting LSTM model and use $M$ to predict $\widehat{v_{t+1}}$;}
7: **else if** $t \geq 2b - 1$ and $t < 2b + 1$ { //i.e., $5 \leq t < 7$, if $b = 3$
8: Calculate $AARE_t$ based on Equation 1;
9: Train an LSTM model by taking $[v_{t-b+1}, v_{t-b+2} \dots, v_t]$ as the training data;
10: Let $M$ be the resulting LSTM model and use $M$ to predict $\widehat{v_{t+1}}$;}
11: **else if** $t \geq 2b + 1$ and *flag*==True { //i.e., $t \geq 7$ if $b = 3$
12: **if** $t \neq 7$ { Use $M$ to predict $\widehat{v_t}$;}
13: Calculate $AARE_t$ based on Equation 1;
14: Calculate $thd$ based on Equation 2;
15: **if** $AARE_t \leq thd$\{ $v_t$ is not considered as an anomaly;}
16: **else**{
17: Train an LSTM model with $[v_{t-b}, v_{t-b+1}, \cdots, v_{t-1}]$;
18: Use the newly trained LSTM model to predict $\widehat{v_t}$;
19: Calculate $AARE_t$ using Equation 1;
20: Calculate $thd$ based on Equation 2;
21: **if** $AARE_t \leq thd$\{ $v_t$ is not considered as an anomaly;}
22: **else** {
23: $v_t$ is reported as an anomaly immediately;
24: Let *flag* be False;}}}
25: **else if** $t \geq 2b + 1$ and *flag*==False {
26: Train an LSTM model with $[v_{t-b}, v_{t-b+1}, \cdots, v_{t-1}]$;
27: Use the newly trained LSTM model to predict $\widehat{v_t}$;
28: Calculate $AARE_t$ based on Equation 1;
29: Calculate $thd$ based on Equation 2;
30: **if** $AARE_t \leq thd$\{
31: $v_t$ is not considered as an anomaly;
32: Replace $M$ with the new LSTM model from line 26;
33: Let *flag* be True;}
34: **else** {
35: $v_t$ is reported as an anomaly immediately; Let *flag* be False;}}}

**Fig. 1.** The algorithm of RePAD.

When $t = 7$ (i.e., $t = 2b + 1$), RePAD calculates $AARE_7 = \frac{1}{3} \cdot \left( \frac{|v_5 - \widehat{v_5}|}{v_5} + \frac{|v_6 - \widehat{v_6}|}{v_6} + \frac{|v_7 - \widehat{v_7}|}{v_7} \right)$. At this moment, RePAD is able to officially detect anomalies since it has $b$ AARE values (i.e., $AARE_5$, $AARE_6$, and $AARE_7$), which are sufficient. In

other words, RePAD requires a preparation period of $2b+1$ time points to get its detection started. Since $b$ is a small integer, the preparation period is very short. Right after calculating $AARE_t$, RePAD immediately uses Equation 2 to calculate $thd$, based on the Three-Sigma Rule [2], which is commonly used for anomaly detection, by considering all previously derived AARE values.

$$thd = \mu_{AARE} + 3 \cdot \sigma, t \geq 2b+1 \quad (2)$$

where $\mu_{AARE} = \frac{1}{t-b-1} \cdot \sum_{x=2b-1}^{t} AARE_x$ (i.e., the average AARE), and $\sigma$ is the standard deviation, which can be derived as below.

$$\sigma = \sqrt{\frac{\sum_{x=2b-1}^{t}(AARE_x - \mu_{AARE})^2}{t-b-1}} \quad (3)$$

If $AARE_7$ is smaller than or equal to $thd$ (see line 15 of Fig. 1), it means that $v_7$ is similar to previous data points. In this case, RePAD does not consider $v_7$ an anomaly, and it keeps using the current LSTM model (i.e., $M$) to predict the next data point $\widehat{v_8}$. However, if $AARE_7$ is larger than $thd$ (see line 16), implying that either the data pattern has changed or an anomaly might happen, RePAD trains an new LSTM model by taking the most recent $b$ data points, i.e., $[v_4, v_5, v_6]$, as the training data. After that, the newly trained LSTM model is then immediately used to re-predict $\widehat{v_7}$ and the corresponding $AARE_7$ is calculated.

If the new $AARE_7$ is smaller than or equal to $thd$ (see line 21), RePAD concludes that the data pattern has slightly changed and that $v_7$ is not an anomaly. This double-check approach (i.e., re-training a new LSTM model and re-comparing the newly calculated $AARE_t$ with $thd$) enables RePAD to adapt to minor pattern changes without being too sensitive.

On the contrary, if the new $AARE_7$ is still larger than $thd$ (see line 22), RePAD considers $v_7$ to be an anomaly since the LSTM model trained with the most recent data points is still unable to correctly predict $v_7$. At this time point, a notification will be immediately sent from RePAD to indicate an anomaly and allow for corresponding actions or countermeasures to be taken.

The above detection process will repeat over and over again as time advances. Based on the above design, it is clear that RePAD only performs LSTM training when needed. This feature is a key factor with respect to why RePAD is able to provide real-time and proactive anomaly detection.

## 5 Experiment Results

To evaluate RePAD, we conducted two experiments, and for each experiment we compared RePAD with the two previously mentioned anomaly detection approaches proposed by Twitter: ADT and ADV. Recall that ADT is designed to detect statistically significant anomalies in a given time series, and that ADV is used to detect statistically significant anomalies in a given vector of observations.

In the first experiment, one of the realAWSCloudwatch datasets called rds-cpu-utilization-e47b3b was chosen from NAB [17] to evaluate the three abovementioned approaches. This dataset is abbreviated as CPU-b3b in this paper. In the second experiment, another time series dataset called Machine Temperature System Failure, abbreviated as MTSF, was chosen from NAB. Table 1 lists the details of the two datasets.

**Table 1.** Two real-world time-series datasets used in the experiments.

| Name | Time Period | # of data points |
|---|---|---|
| CPU-b3b | From 2014-04-10, 00:02 to 2014-04-23, 23:57 | 4032 |
| MTSF | From 2013-12-02, 21:15 to 2014-02-19, 15:25 | 22695 |

Note that the interval time between data points in both datasets is 5 minutes. Hence, all the three approaches also followed the same time interval to conduct their detection. Recall that the Look-Back parameter and Predict-Forward parameter of RePAD should be small integers. Hence, we set $b$ as 3 and $f$ as 1 in both experiments. On the other hand, we were required to set parameter $k$ for both ADT and ADV. Note that parameter $k$ indicates the maximum number of anomalies to detect, represented as a percentage of the whole dataset [18]. In both experiments, $k$ is 0.02 based on the setting mentioned in [15]. Furthermore, we had to set the *period* parameter for ADV. We set this parameter to be 1440 by following the setting in [15]. All experiments are run on a laptop running MacOS 10.15.1 with 2.6 GHz 6-Core Intel Core i7 and 16GB DDR4 SDRAM. In this paper, we do not use the traditional metrics such as true positive, false positive, true negative, and false negative to measure the prediction performance of the three approaches, since these metrics are unable to effectively test anomaly detection algorithms for real-time uses [1]. Instead, we evaluate how early each approach is able to detect an anomaly before the occurrence of the anomaly since the goal of RePAD is to offer real-time proactive anomaly detection for time series.

### 5.1 Experiment 1

Fig. 2 illustrates all the data points in the CPU-b3b dataset and the detection results of RePAD, ADT, and ADV on this dataset. In this dataset, there are two anomalies labeled by human experts, and both of them are marked as red circles in the figure. When RePAD was employed, it made two false anomaly detections in the preparation period. But this situation did not last in the period between the beginning of the time series and the first anomaly since RePAD has learned the data pattern. Figs. 3 and 4 show a close-up of the detection results on the first and second anomalies, respectively. It is clear that RePAD is the only approach that is able to detect these two anomalies on time. Both ADT and ADV are unable to detect these two anomalies.

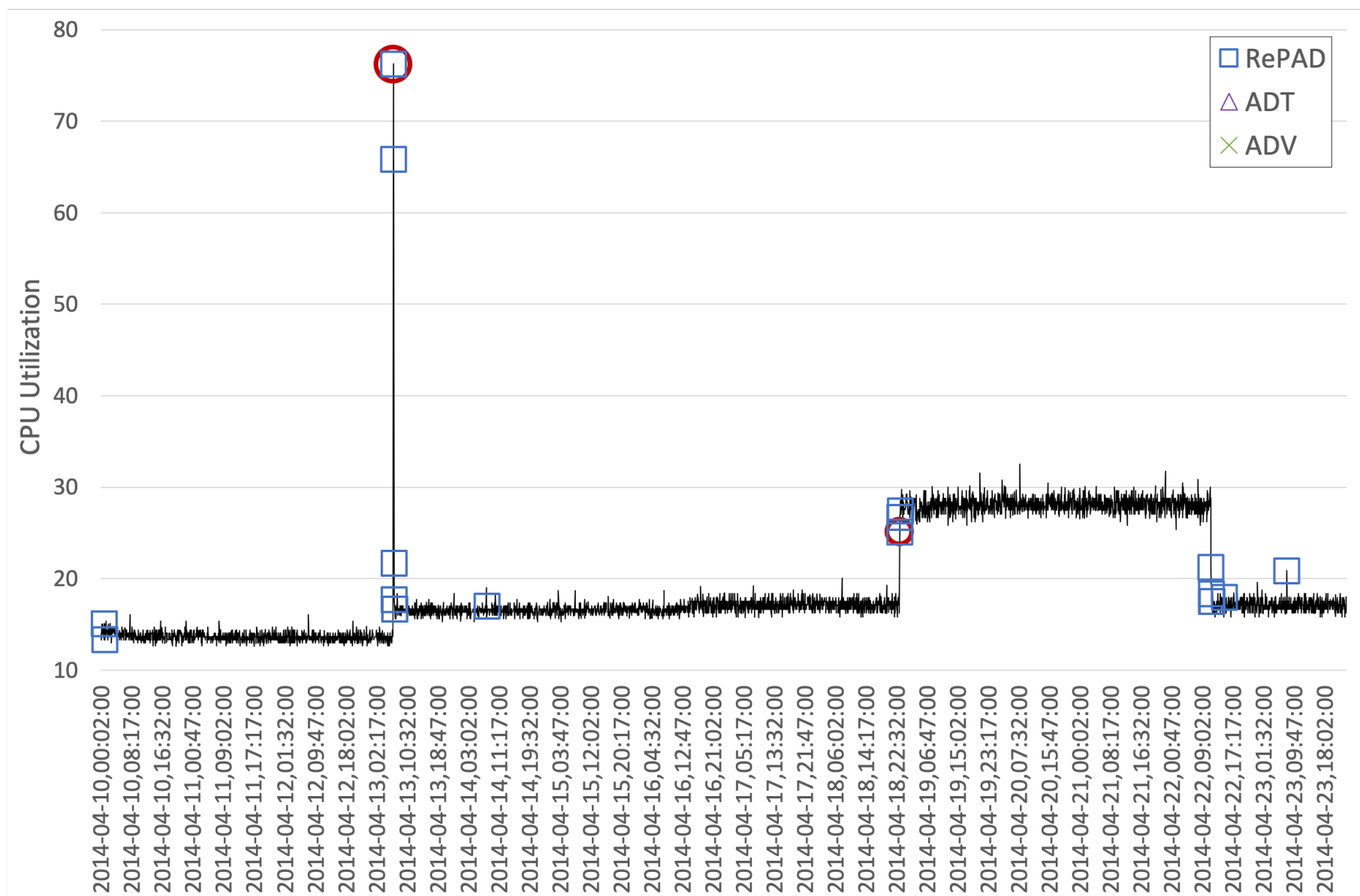

**Fig. 2.** The detection results of RePAD, ADT, and ADV on the CPU-b3b dataset. Note that this dataset has two anomalies labeled by human experts, and they are marked as red circles.

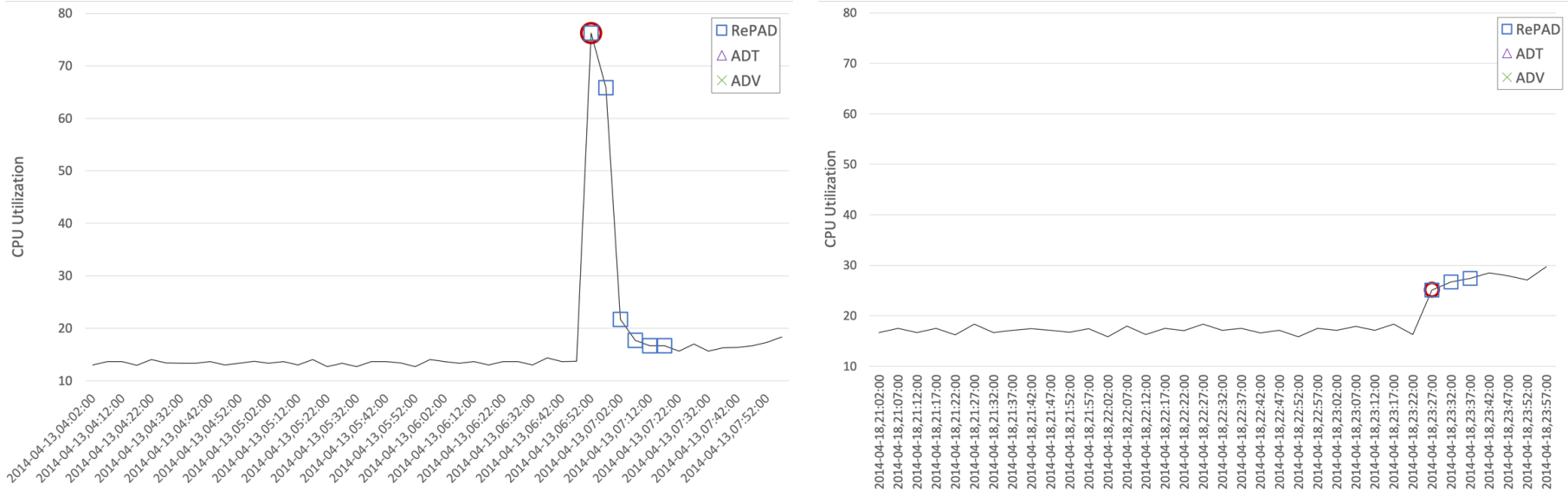

**Fig. 3.** The close-up of the detection results for the first anomaly in the CPU-b3b dataset.

**Fig. 4.** The close-up of the detection results for the second anomaly in the CPU-b3b dataset.

Table 2 lists the time performance of RePAD on detecting anomalies in the CPU-b3b dataset. RePAD needs to perform LSTM retraining 38 times. We can see that the LSTM retraining ratio is very low since it is only 0.94% (=38/4027). Note that the LSTM retraining ratio is the ratio between the total number of LSTM retraining required by RePAD and the total number of LSTM retraining in the worst case (which means that RePAD has to retrain an LSTM model at every time point after the preparation period). In addition, as listed in Table 2, the average time required by RePAD to make a detection decision for each data point in the CPU-b3b dataset is only 0.022 second with a low standard deviation, implying that RePAD is able to provide real-time anomaly detection and offer real-time LSTM retraining in real time. Note that the above measures are inapplicable to ADT and ADV since both of them are statistical-based approaches rather than real-time approaches.

**Table 2.** The time performance of RePAD on the CPU-b3b dataset.

| # of data points that requires LSTM retraining | 38 |
|---|---|
| LSTM retraining ratio | 0.94% (=38/4027) |
| Average Detection Time (sec) | 0.022 |
| Standard deviation (sec) | 0.033 |

### 5.2 Experiment 2

Fig. 5 illustrates all the data points in the MTSF dataset and the corresponding detection results of the three approaches. In this dataset, there are three anomalies labeled by human experts, and all of them are marked as red circles in Fig. 5. Note that the red diamond shown in Fig. 5 is labeled by human experts as a sign leading to the last anomaly. We can see that RePAD makes a number of false warnings in the beginning since the pattern in this dataset experiences significant deviation. However, we can see that between the second anomaly and the sign (i.e., the red diamond), RePAD did not make any false prediction since RePAD has learned the pattern of this dataset.

Fig. 6 shows a close-up of the detection result on the first anomaly. Apparently, only RePAD is able to proactively detect this anomaly – around 450 minutes earlier than the occurrence of the anomaly. On the other hand, we can see that neither ADT nor ADV is able to detect this anomaly.

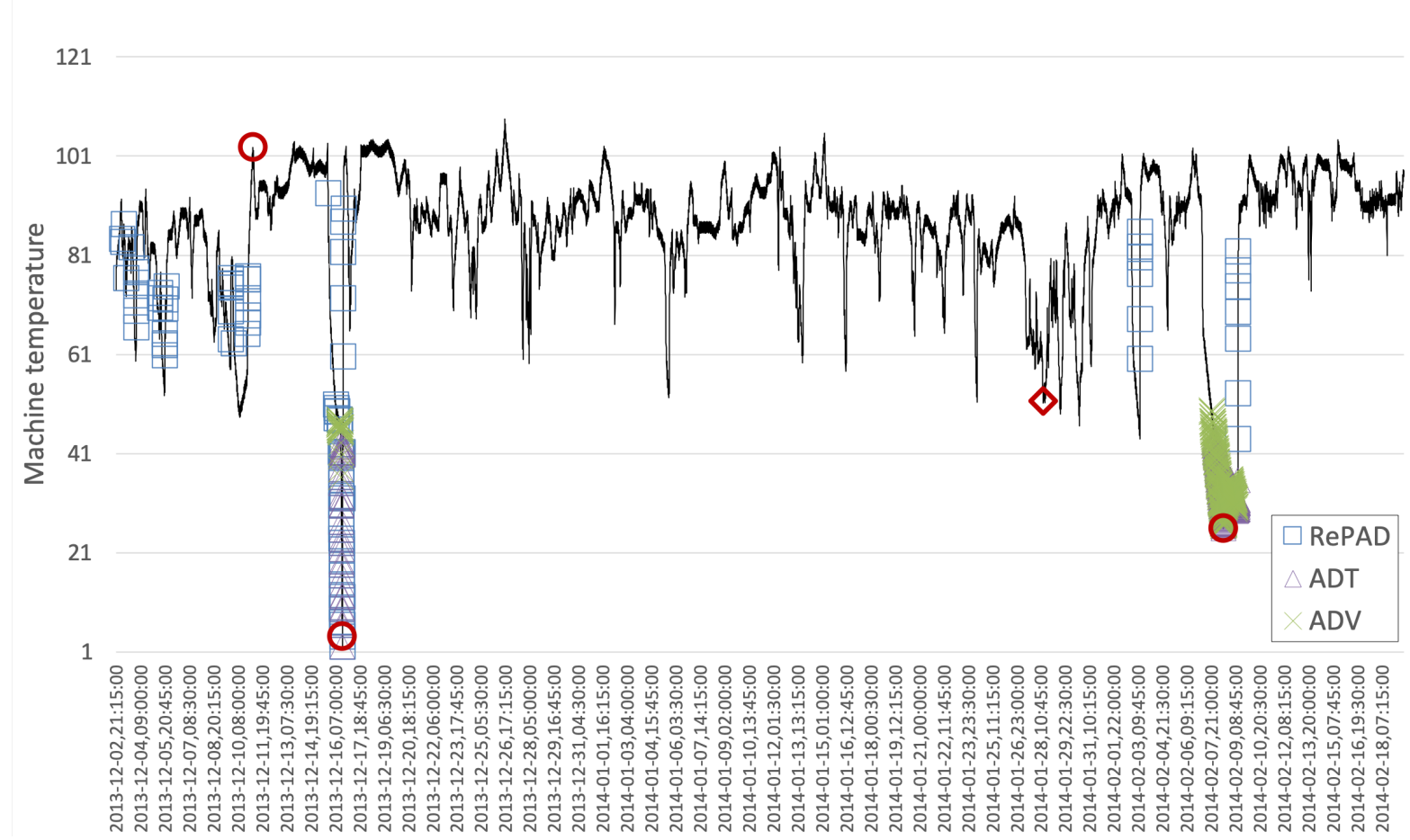

**Fig. 5.** The detection results of RePAD, ADT, and ADV on the MTSF dataset. Note that this dataset has three anomalies labeled by human experts, marked as red circles. In addition, there is one sign labeled by the human experts before the last anomaly. This sign is marked as a red diamond.

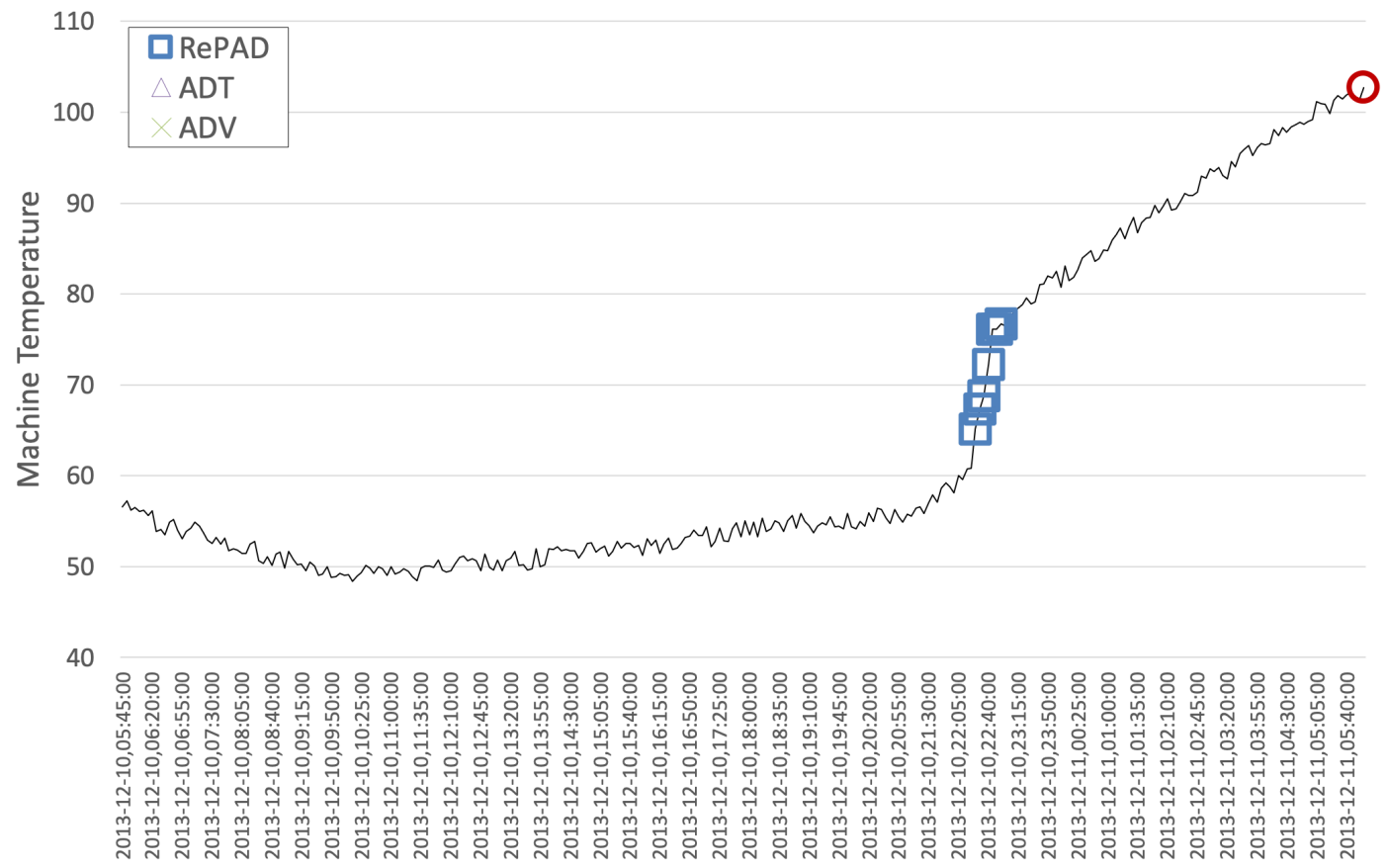

**Fig. 6.** The close-up of the detection results for the first anomaly in the MTSF dataset.

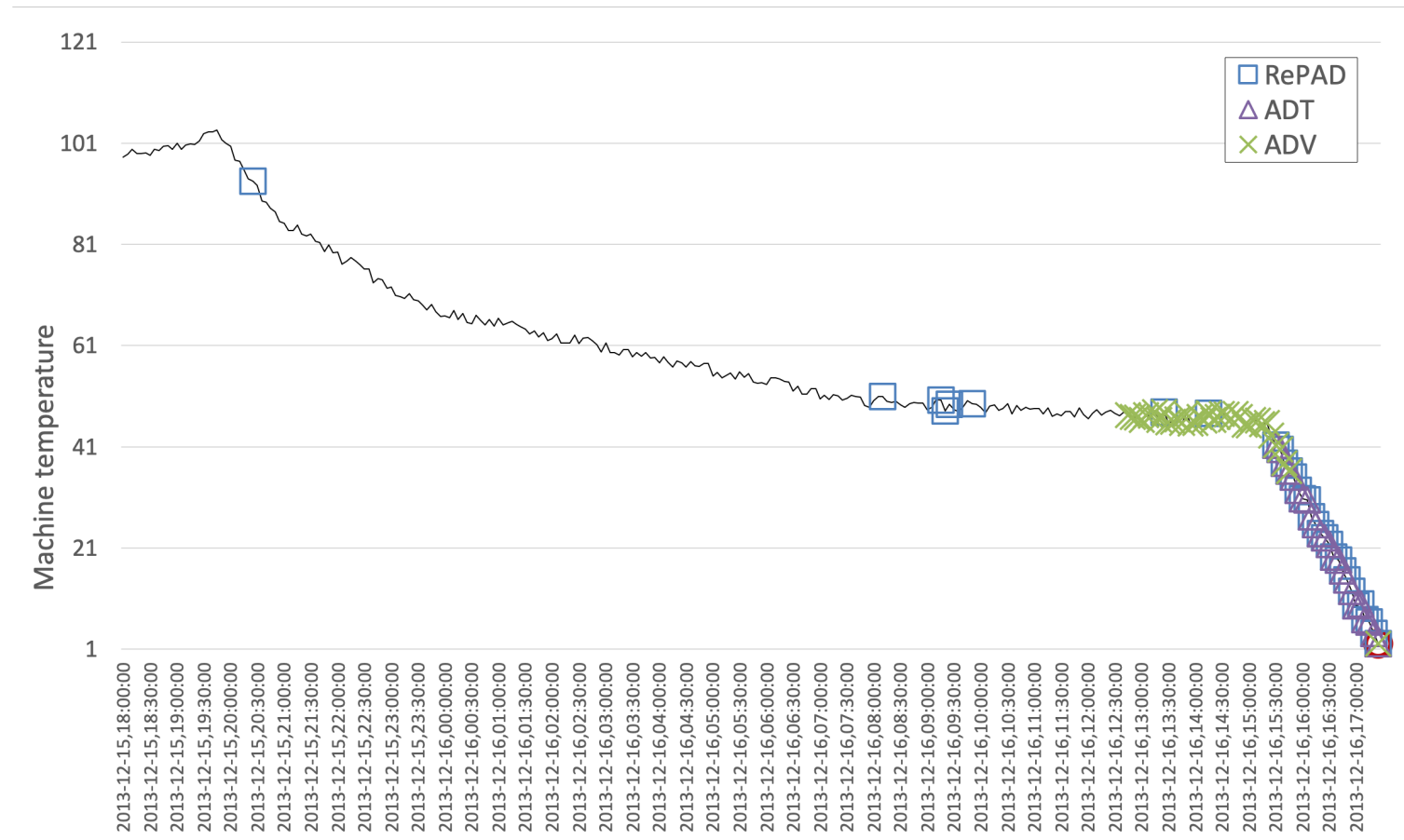

**Fig. 7.** The close-up of the detection results for the second anomaly in the MTSF dataset.

Fig. 7 depicts a close-up of the detection results on the second anomaly in the MTSF dataset. All the three approaches are able to detect this anomaly, but RePAD outperforms ADT and ADV since it is able to detect this anomaly much earlier than ADT and ADV. The time difference between the first anomaly notification sent by RePAD and the anomaly is 1255 minutes (around 20.92 hours). This long period should provide sufficient time for system administrators or any other responsible persons to take appropriate actions.

The superior detection performance of RePAD can also be seen from the detection of the last anomaly. From Fig. 8, we see that after the sign of the last anomaly shows up, RePAD is the first to proactively detect the last anomaly.

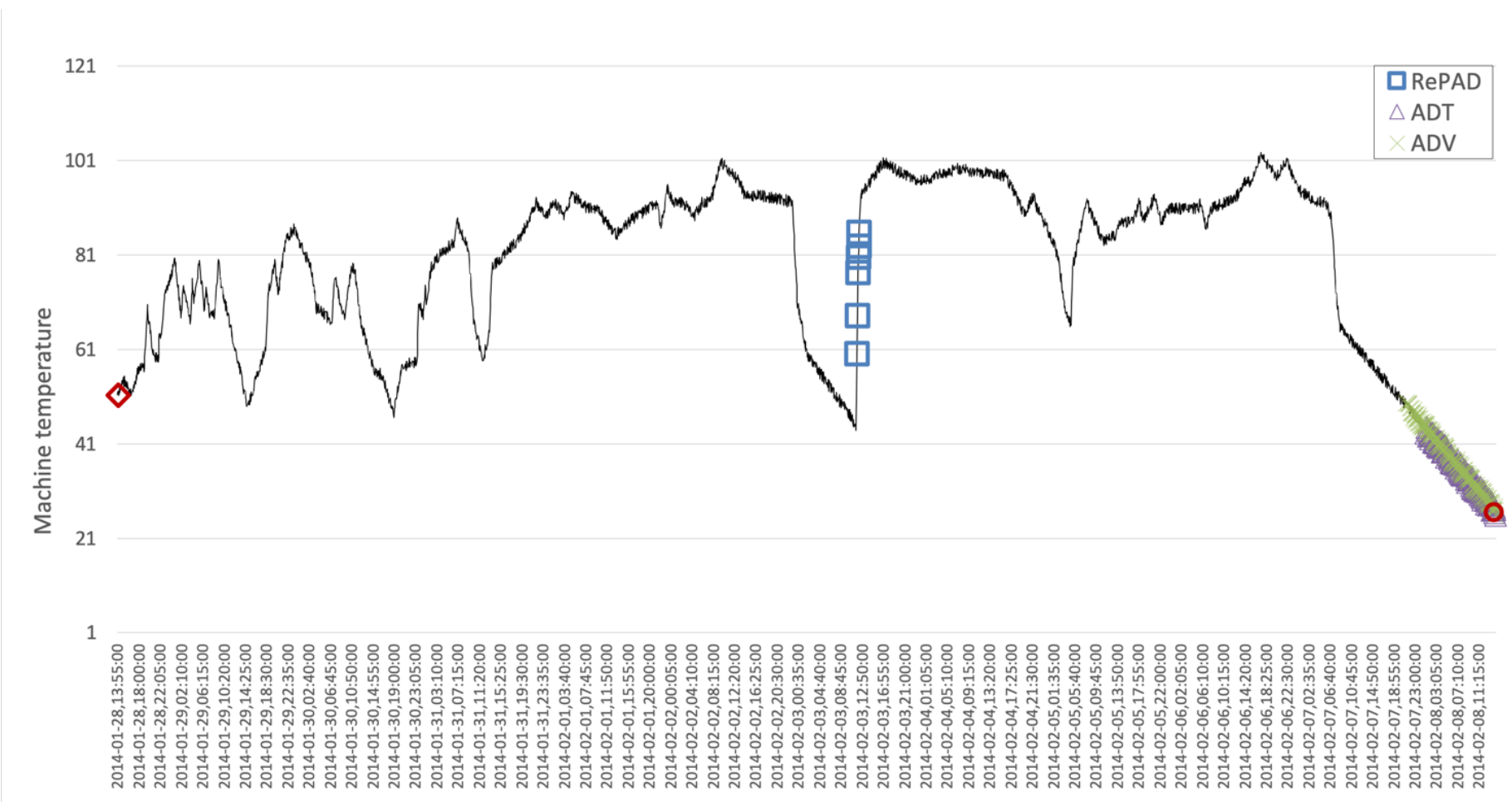

**Fig. 8.** The close-up detection results for the third anomaly in the MTSF dataset.

Table 3 lists the time performance of RePAD on the MTSF dataset. RePAD needs to conduct LSTM retraining 134 times, but the LSTM retraining ratio is still very low. It is only 0.59% (=134/22690). Furthermore, the average detection time required by RePAD is only 0.283 second with a low standard deviation, demonstrating the real-time anomaly detection capability of RePAD.

**Table 3.** The time performance of RePAD on the MTSF dataset.

| # of data points that requires LSTM retraining | 134 |
|---|---|
| LSTM retraining ratio | 0.59% (=134/22690) |
| Average Detection Time (sec) | 0.283 |
| Standard deviation (sec) | 0.271 |

## 6 Conclusion and Future Work

In this paper, we have introduced RePAD, an algorithm that is able to proactively detects anomalies in time series in real time. RePAD is able to work on any time series without pre-collecting training data and going through an off-line training process. After a very short preparation period, RePAD start its detection function, retrains an LSTM model when necessary, and dynamically adjusts its detection threshold over time. These features enable RePAD to adapt to pattern changes in time series and proactively detect anomalies in a time-efficient way. Experiments based on real-world time series datasets demonstrate that RePAD provides a satisfactory detection performance since it can detect all the given anomalies either proactively or on time.

As future work, we plan to further improve RePAD on reducing its false warnings/anomalies using a hybrid approach based on multi-LSTMs. In addition, we would like to extend RePAD for large-scale time series data on the eX[3] HPC cluster [21] by referring to [19][20] and designing it in a parallel way.

**Acknowledgments.** This work was supported by the project eX$^3$ - *Experimental Infrastructure for Exploration of Exascale Computing* funded by the Research Council of Norway under contract 270053 and the scholarship under project number 80430060 supported by Norwegian University of Science and Technology.